\documentclass[letterpaper, 10 pt, conference]{ieeeconf}
\DeclareUnicodeCharacter{0301}{\'{e}}
\IEEEoverridecommandlockouts
\usepackage{graphicx}
\usepackage{tabularx}
\usepackage{subcaption}
\usepackage[linesnumbered, ruled,vlined]{algorithm2e}
\usepackage{array}
\usepackage[caption=false,font=normalsize,labelfont=sf,textfont=sf]{subfig}
\usepackage{textcomp}
\usepackage{stfloats}
\usepackage{url}
\usepackage{verbatim}
\usepackage[table,xcdraw]{xcolor} %
\usepackage{colortbl}
\usepackage[caption = false]{subfig}
\usepackage{graphicx} %
\usepackage{caption}
\usepackage{stfloats} %
\usepackage{tabularx}
\usepackage{cite}
\usepackage{xspace}
\newcommand{\ari}{{\textit{{MBE-ARI}}}\xspace}
\usepackage[scaled=.7]{beramono}

\usepackage{microtype}
\usepackage{amsmath} %
\usepackage{amssymb}  %
\usepackage{booktabs}
\usepackage{balance}
\usepackage{float}
\usepackage{multirow}
\usepackage{xcolor}
\usepackage{algpseudocode}
\usepackage{pifont}
\begin{document}

\title{\textit{MBE-ARI}: A Multimodal Dataset Mapping Bi-directional\\ Engagement in Animal-Robot Interaction}
\author{Ian Noronha$^{1}$*, Advait Prasad Jawaji$^{2}$*, Juan Camilo Soto$^{1}$*, Jiajun An$^{1}$, Yan Gu$^{2}$, and Upinder Kaur$^{1\dagger}$%
\thanks{*These authors contributed equally.$^\dagger$Corresponding author. The authors acknowledge the support of the Purdue Ag-ENG seed grant in the fulfillment of this work. The authors acknowledge the support of Dr. Hinayah Rojas de Oliviera and Dr. Heather Neave in the design and execution of the data collection studies.}%
\thanks{$^{1}$Ian Noronha, Juan Camilo Soto, Jiajun An, and Upinder Kaur are with the Department of Agricultural and Biological Engineering, Purdue University, 401 Grant Street, West Lafayette, Indiana, USA.
        {\tt\footnotesize \{inoronha, soto97, an80,  kauru\}@purdue.edu.}}%
\thanks{$^{2}$Advait Prasad Jawaji and Yan Gu are with the School of Mechanical Engineering, Purdue University, 585 Purdue Mall, West Lafayette, Indiana, USA.
        {\tt\footnotesize \{ajawaji, yangu\}@purdue.edu.}}%
}

\maketitle
\begin{center}
    \textit{Accepted to ICRA 2025}
\end{center}
\thispagestyle{empty}
\pagestyle{plain}

\begin{abstract}
Animal-robot interaction (ARI) remains an unexplored challenge in robotics, as robots struggle to interpret the complex, multimodal communication cues of animals, such as body language, movement, and vocalizations. Unlike human-robot interaction, which benefits from established datasets and frameworks, animal-robot interaction lacks the foundational resources needed to facilitate meaningful bidirectional communication. To bridge this gap, we present the \textit{MBE-ARI} (Multimodal Bidirectional Engagement in Animal-Robot Interaction), a novel multimodal dataset that captures detailed interactions between a legged robot and cows. The dataset includes synchronized RGB-D streams from multiple viewpoints, annotated with body pose and activity labels across interaction phases, offering an unprecedented level of detail for ARI research. Additionally, we introduce a full-body pose estimation model tailored for quadruped animals, capable of tracking 39 keypoints with a mean average precision (mAP) of 92.7\%, outperforming existing benchmarks in animal pose estimation. The MBE-ARI dataset and our pose estimation framework lay a robust foundation for advancing research in animal-robot interaction, providing essential tools for developing perception, reasoning, and interaction frameworks needed for effective collaboration between robots and animals. The dataset and resources are publicly available at \url{https://github.com/RISELabPurdue/MBE-ARI/}, inviting further exploration and development in this critical area. 
\end{abstract}

\section{Introduction}
In recent years, robotics has made significant strides in mimicking animal forms and behaviors. Innovations in adaptive grasping~\cite{ruotolo2021grasping}, agile locomotion~\cite{legged_loco,baisch2014high}, and biomimetic designs~\cite{agile_robo_fish,robotic_hand} have empowered robots to operate effectively across various environments. Yet, despite these advances, robots still struggle to communicate and interact meaningfully with sentient animals--creatures capable of perception, complex behaviors, and conscious experience. This gap in animal-robot interaction (ARI) presents a critical challenge to progress in essential fields such as animal welfare, environmental conservation, and scientific understanding of interspecies interactions.

\begin{figure}[!ht]
\vspace{0.6em}
\centering\includegraphics[width=1.0\linewidth]{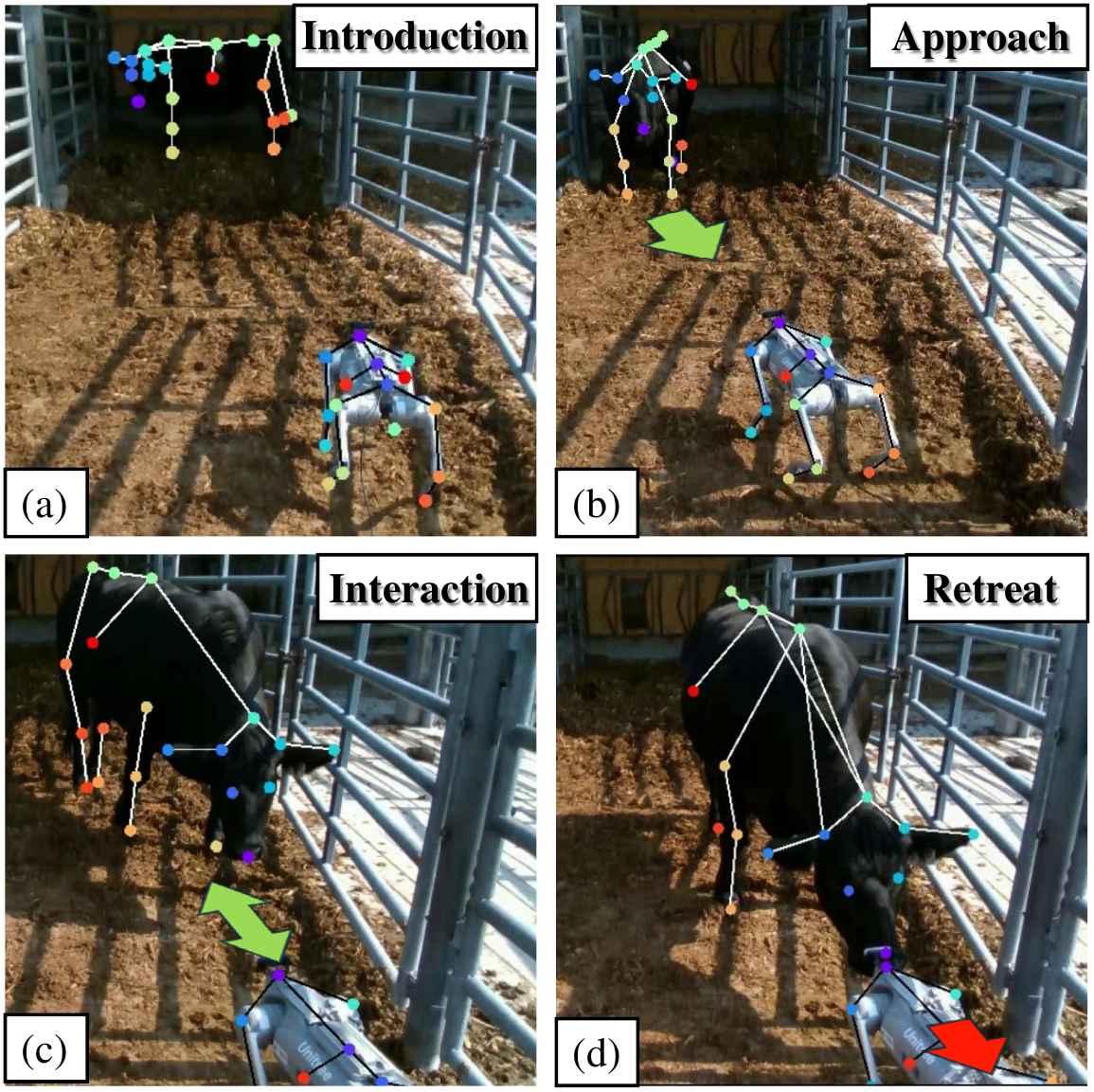}
\caption{Example frames from the \ari dataset with pose estimation, illustrating key stages of animal-robot interaction: (a) Introducing the cow to the static robot; (b) Approach as the cow and robot move towards each other; (c) Interaction between the cow and robot; and (d) Retreat stage as the robot withdraws when the cow gets too close.}
\vspace{-1.25em}
\label{fig:intro}
\end{figure}

ARI is an emerging field poised to unlock new insights in robotics by enabling robots to engage meaningfully with animals that exhibit complex behaviors and decision-making. By studying these interactions, we can develop advanced perception, learning, and adaptive control systems. While previous research has used robots to observe simpler organisms such as zebrafish and rats, capable of perception but with limited cognition and conscious experiences, these studies provide a limited understanding of the rich dynamics between robots and more complex animals. For instance, robotic zebrafish have been used to study changes in group behavior~\cite{zebra_fish, zebra_fish_12}, while research with rats has revealed key aspects of spatial memory and avoidance learning in response to robotic threats~\cite{rat_behave_telensky, rat_ahuja}. However, the cognitive limitations of these organisms restrict our understanding of nuanced interspecies communication, which is more prevalent in sentient mammals such as cattle and other complex wildlife. 

Shifting our focus to interactions with more complex, sentient animals, allows us to push the boundaries of current robotics capabilities. These interactions present unique challenges, as robots must interpret subtle and multimodal behavioral cues, similar to the complexities found in human-robot interaction (HRI).  However, unlike HRI, where established frameworks enable robots to interpret human language and gestures~\cite{speech_hri,language_social_robotics}, there are no comparable systems for decoding animal communication. For instance, an animal's subtle eye and ear movements can signal fear, curiosity, or discomfort~\cite{lambert2019positive}-- cues that robots currently struggle to comprehend. Bridging this gap requires a new research paradigm, as existing robotic systems are ill-equipped to interpret the sophisticated and non-verbal behaviors exhibited by sentient animals.

In this work, we address the critical gap in ARI by introducing the \ari (Multimodal Bidirectional Engagement in Animal-Robot Interaction) dataset--a novel multimodal dataset that captures detailed interactions between a legged robot and a sentient mammal: the cow. Cows, like humans, are sentient social beings whose health and well-being are closely related to everyday social interactions. However, their stoic nature can obscure signs of distress or discomfort, complicating their care in large commercial facilities, where rapid and accurate assessments are crucial. With global concerns over food security, climate change, and sustainability, improving cow welfare through precise and responsive robotic systems is more important than ever~\cite{kaur2021future, kaur2023invited}. By focusing on this complex animal, the \textit{MBE-ARI} dataset lays a foundation for advancing ARI research, providing the tools needed for robots to better understand and respond to animal behaviors. 

Our approach draws on insights from social robotics frameworks, which have been successful in reducing the perceived threat of robots to humans and fostering bidirectional communication through gestures and language~\cite{language_social_robotics,robot_gaze,olaronke_robot_cues}. Adapting these principles to ARI, we designed experiments to facilitate meaningful engagement between robots and cows. In particular, we identified critical behavioral cues--such as ear and eye movements, body posture, and gait--that signal a cow's emotional state and guide their interactions~\cite{lambert2019positive}. Capturing these cues with multiple camera viewpoints and depth measurements, we created the MBE-ARI dataset to provide a comprehensive resource for analyzing these interactions. Each experiment is meticulously labeled, categorizing tasks, interaction goals, and success or failure, ensuring that this structured dataset will serve as a valuable foundation for future research in the field. %

While social robotics has advanced significantly with AI-driven frameworks for perception, reasoning, and planning, similar tools are missing in the domain of ARI. To address this gap, we developed a novel full-body pose estimation system, leveraging the rich data from the MBE-ARI dataset. Our system is fine-tuned to the unique physiology of cows, tracking 39 critical keypoints across the body that help identify subtle behavioral cues during interactions, as shown in Fig.~\ref{fig:intro}. By capturing these detailed signals, such as body posture and eye movements, our framework provides new insights into animal responses. Additionally, we implemented a companion model to track the robot’s pose during interactions, ensuring precise monitoring of its movements relative to the animal. Together, these developments establish a new benchmark for research in animal-robot interaction, laying the groundwork for autonomous robotic systems that can enhance animal welfare in real-world environments.

\begin{figure}[!tb]
\vspace{1.25em}
\centering\includegraphics[width=0.98\linewidth]{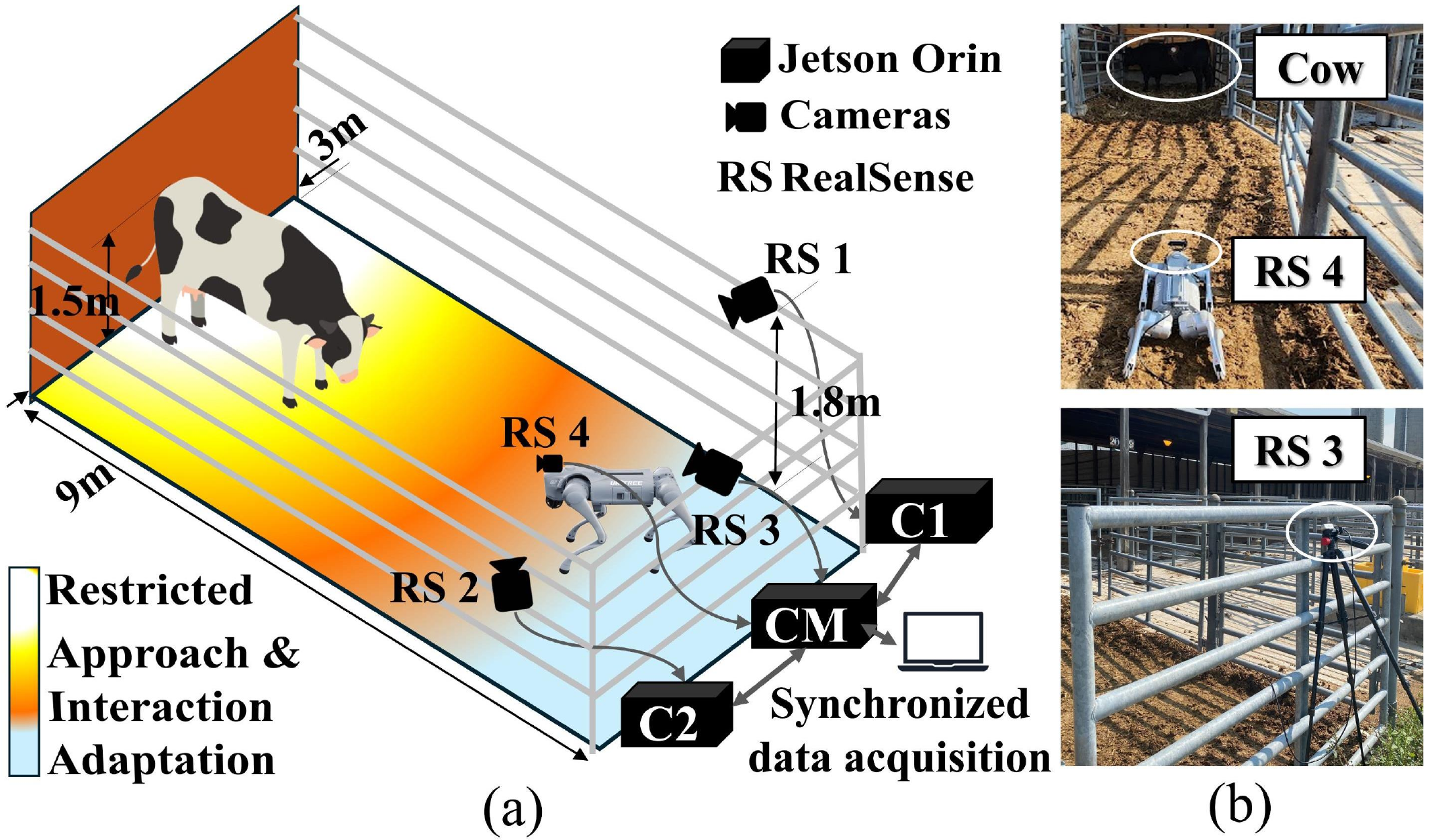}
\caption{(a) The collection setup showing key dimensions and arrangement of the capturing technology. (b) Images of the robot and camera setup.}
\vspace{-1.25em}
\label{fig:setup}
\end{figure}

To this end, our contributions are:
\begin{itemize}
    \item A novel multimodal dataset--\ari-- capturing interactions between a legged robot and a sentient animal (cow). This dataset provides labeled and synchronized multimodal data, establishing a foundational resource for future research in ARI. 
    \item A structured task-based annotation structure that identifies and labels key interaction phases and nuances of ARI. This framework enables researchers to develop more effective bidirectional interaction strategies and analyze complex behavioral dynamics. 
    \item A full-body pose estimation system designed for quadruped animals, tracking 39 keypoints, significantly advancing the accuracy of interaction cue assessment. This system sets a new benchmark for animal pose estimation frameworks
    \item Extensive validation of the pose estimation framework on the \ari dataset, demonstrating its effectiveness when compared to existing benchmarks.
\end{itemize}

\section{\ari Dataset}
The MBE-ARI dataset offers a unique and comprehensive collection of multimodal data, capturing the critical stages of interaction between a legged robot and a sentient mammal—cows. Through carefully designed experiments, the dataset captures rich behavioral and interactional data, offering synchronized RGB-D streams, body pose, and environmental context. The data collected provides invaluable insight into the nuanced dynamics of animal-robot interaction, addressing key gaps in the field.

\subsection{Experiment Design}
Our experimental design was developed to replicate real-world animal-robot interactions in a controlled environment while ensuring ethical standards were strictly followed. All experiments adhered to the IACUC guidelines (protocol IACUC 0324002489), and trained animal science professionals were present to handle the cows, minimizing any external influence on natural interactions.

To study how a legged robot interacts with cows, we designed experiments that mimic the natural dynamics between cows and shepherd dogs. These interactions rely on non-verbal communication cues such as gait and posture, which we aimed to replicate. Our controlled experimental setup enabled us to systematically observe how cows respond to the presence and actions of a robotic system.

The experiments were conducted in a secured pen at the Purdue Animal Science Research and Education Center (ASREC) Beef Unit, a state-of-the-art facility for animal behavior studies. We selected two 23-month-old female Angus Beef Heifers each weighing approximately 1300 lbs, for the experiments. To minimize stress and ensure natural behavior, the cows spent two weeks acclimating to the pen and the researchers, ensuring that any observed interactions were not influenced by unfamiliarity with the environment. 

The Unitree Go2 robot was selected for its advanced capabilities in autonomous navigation and dynamic locomotion, ideal for studying ARIs. With 12 degrees of freedom across its four articulated legs, it can navigate uneven terrain and mimic animal-like movement patterns. These features allowed us to study how its proximity, gait, and behavior influenced cow movement. The robot’s onboard sensors, including RGB-D cameras, captured detailed interaction data, providing key insights into both the robot’s and the cows' behaviors.

The experiments were conducted in the afternoon, four hours after feeding, to ensure that the cows were not preoccupied with food. The same team of researchers worked with the cows throughout all sessions, maintaining a consistent environment and minimizing external distractions.

To ensure comprehensive coverage, three Intel RealSense D435i cameras were strategically placed around the pen to capture synchronized RGB-D data without occlusions, as shown in Fig.~\ref{fig:setup}(a)-(b). An additional camera on the robot provided a unique point-of-view perspective of the interactions, mirroring the perspective robots will have in real-world environments. All cameras were calibrated using AprilTags (8.8cm tags on a 26x26in board), ensuring precise alignment for extrinsic meshing and depth analysis.

\subsection{Data Acquisition}

All camera systems operated in a master-slave configuration, ensuring global time synchronization across the dataset. This synchronization was key for correlating events between different viewpoints and maintaining temporal coherence. Furthermore, we recorded the joint angles of the robot and data from the onboard IMU sensor, allowing deeper analysis of both robot actions and cow reactions.

\begin{figure}[!t]
\vspace{0.25em}
    \centering\includegraphics[width=0.95\linewidth]{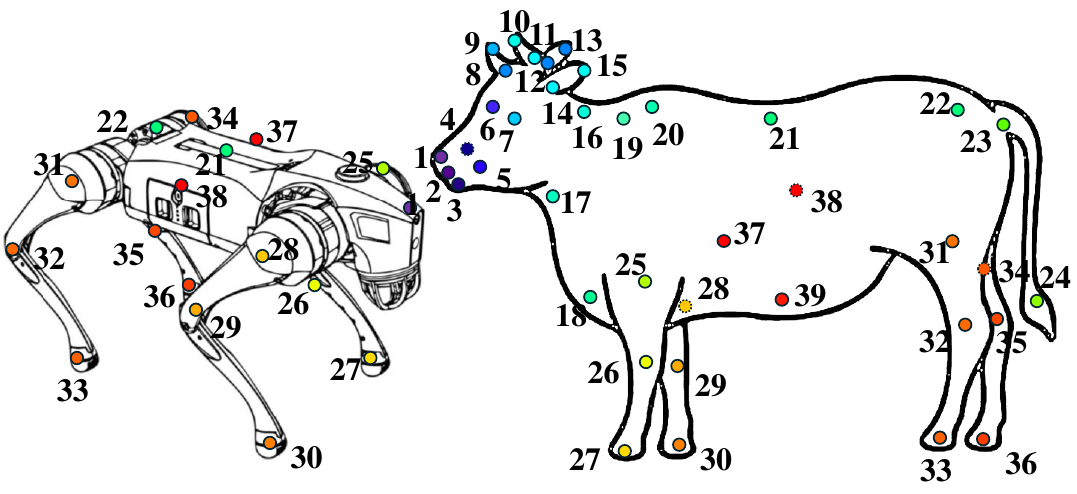}
\small
\begin{tabular}{p{0.015\linewidth}p{0.17\linewidth}p{0.015\linewidth}p{0.18\linewidth}p{0.015\linewidth}p{0.198\linewidth}}
    \toprule
    \textbf{\#} & \textbf{Label}          & \textbf{\#} & \textbf{Label}        & \textbf{\#} & \textbf{Label}         \\
    \midrule
    1  & Nose                & 14 & L Ear Base     & 27 & F L Paw           \\
    2  & Upper Jaw           & 15 & L Ear Tip      & 28 & F R Thigh         \\
    3  & Lower Jaw           & 16 & Neck Base      & 29 & F R Knee          \\
    4  & R M End         & 17 & Throat Base    & 30 & F R Paw           \\
    5  & L M End         & 18 & Throat End     & 31 & H L Thigh         \\
    6  & R Eye               & 19 & Neck End       & 32 & H L Knee          \\
    7  & L Eye               & 20 & Back Base      & 33 & H L Paw           \\
    8  & R A Base       & 21 & Back Mid    & 34 & H R Thigh         \\
    9  & R A Tip        & 22 & Back End       & 35 & H R Knee          \\
    10 & L A Tip        & 23 & Tail Base      & 36 & H R Paw           \\
    11 & L A Base       & 24 & Tail End       & 37 & Torso L           \\
    12 & R Ear Base          & 25 & F L Thigh      & 38 & Torso R           \\
    13 & R Ear Tip           & 26 & F L Knee       & 39 & Torso Bottom      \\
    \bottomrule
    \vspace{0.15 cm}
\end{tabular}
\\
\footnotesize{\textbf{Note:} L = Left, R = Right, F = Front, H = Hind, A = Antler, M=Mouth}
\caption{The 39 data body reference points and their description in animals and robots labeled in the \ari dataset.}
\vspace{-0.3cm}
\label{Fig:body_points}
\end{figure}

\subsection{Body Pose Annotations}
The full body pose of the animal conveys essential cues about its response and comfort with the robot. To assess the cows’ comfort and behavioral responses during interactions, we developed detailed annotations of the body pose for both the animals and the robot. Tracking the full-body pose was crucial not only for understanding the animals’ subtle behavioral cues but also for monitoring the robot’s path planning and localization during interactions.

For this purpose, we manually annotated approximately 3,000 frames from the dataset, capturing 39 key body points on both the cow and the robot (illustrated in Fig.~\ref{Fig:body_points}). These keypoints serve as the foundation for our pose estimation model, providing ground truth for accurately tracking interactions. The selected points are based on established literature in animal behavior, physiology, and body pose modeling~\cite{mathis2018deeplabcut,mathis2020deep,lauer2022multi}, ensuring that our annotations were relevant for behavioral analysis.

\subsection{Task Design and Annotations}
\begin{figure*}[!thbp]
\vspace{1.4em}
\centering
\begin{tabular}{ccccc}
    \multicolumn{1}{c}{\textbf{}} &
    \multicolumn{1}{c}{\textbf{Adaptation}} & 
    \multicolumn{1}{c}{\textbf{Approach}} & 
    \multicolumn{1}{c}{\textbf{Interaction}} & 
    \multicolumn{1}{c}{\textbf{Retreat}}\\
    \raisebox{1\height}{\rotatebox[origin=c]{90}{
    \begin{tabular}{c}
    \textit{Realsense D435i}\\
        \textit{848x480 RGB}
    \end{tabular}}
} &
    {\includegraphics[width=0.2\linewidth]{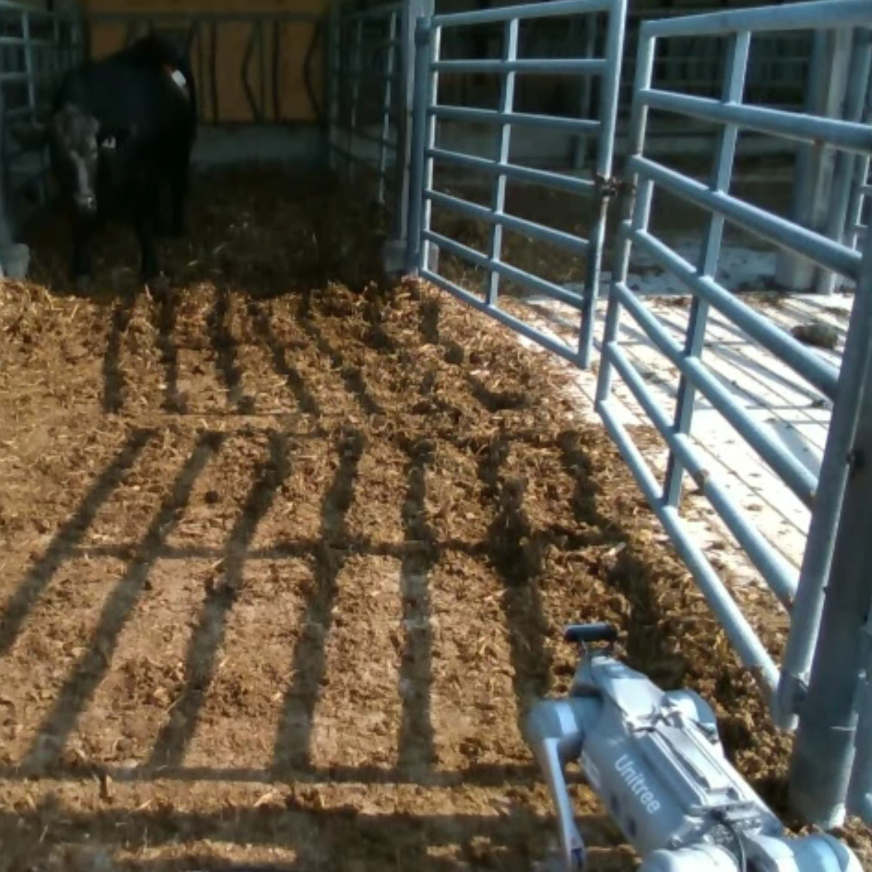} }  &
    {\includegraphics[width=0.2\linewidth]{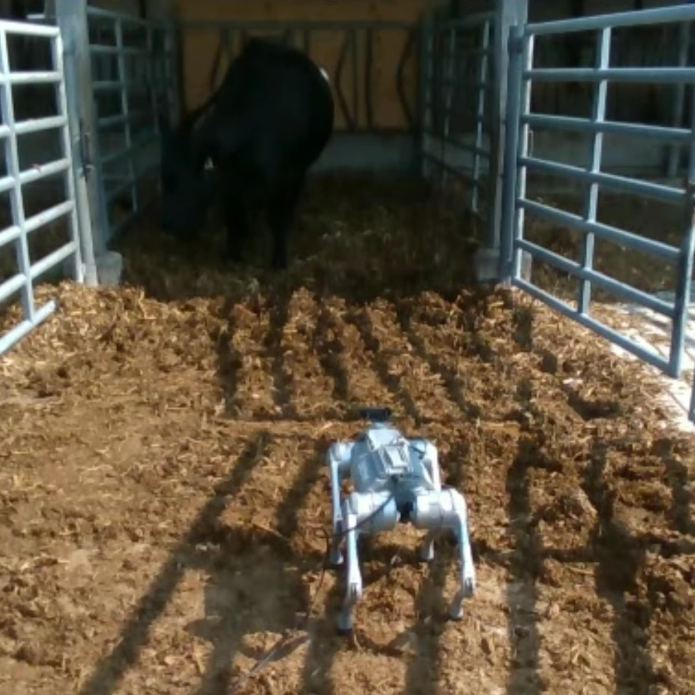}}   &
    {\includegraphics[width=0.2\linewidth]{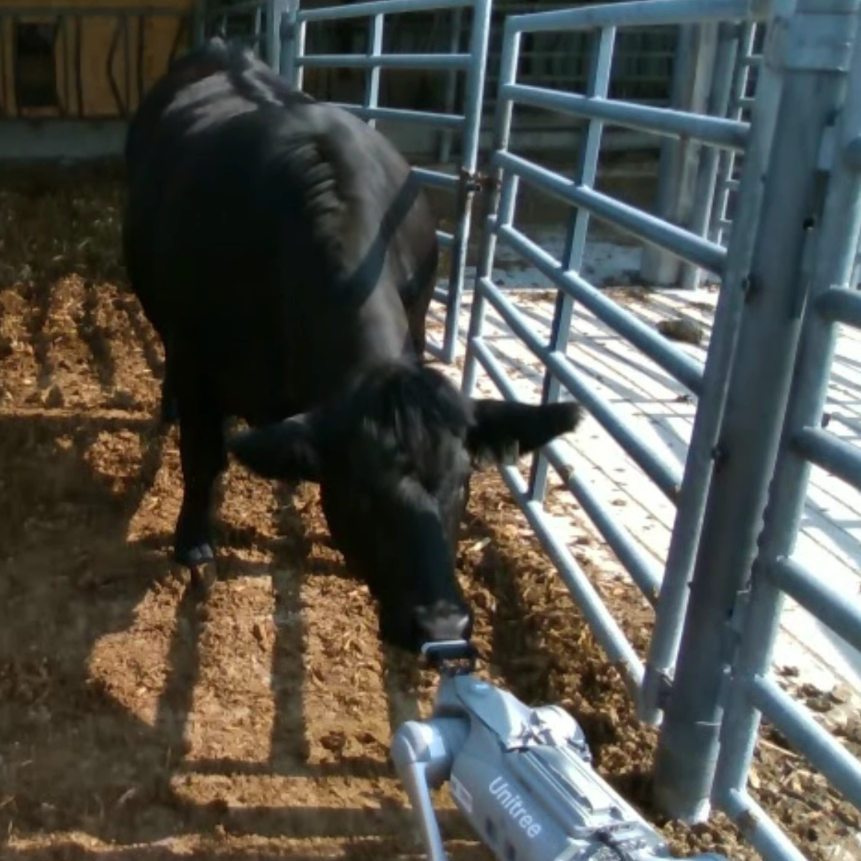}}   &
    {\includegraphics[width=0.2\linewidth]{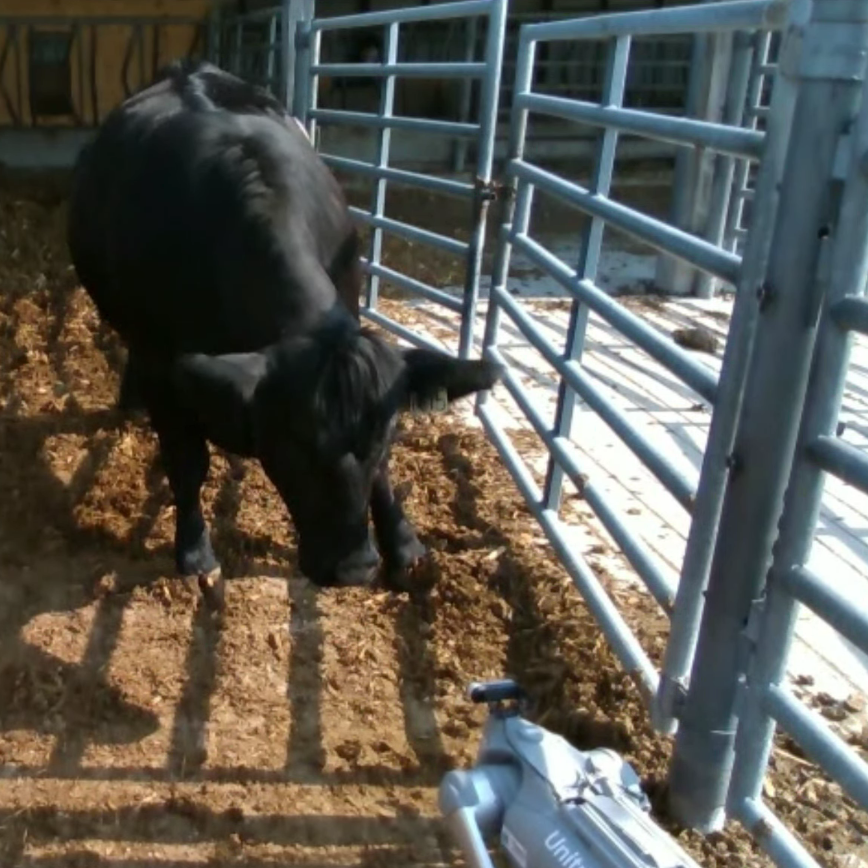}}  \\
    
\raisebox{1\height}{\rotatebox[origin=c]{90}{
    \begin{tabular}{c}
    \textit{Realsense D435i}\\
        \textit{848x480 Depth}
        
    \end{tabular}}
} &
    {\includegraphics[width=0.2\linewidth]{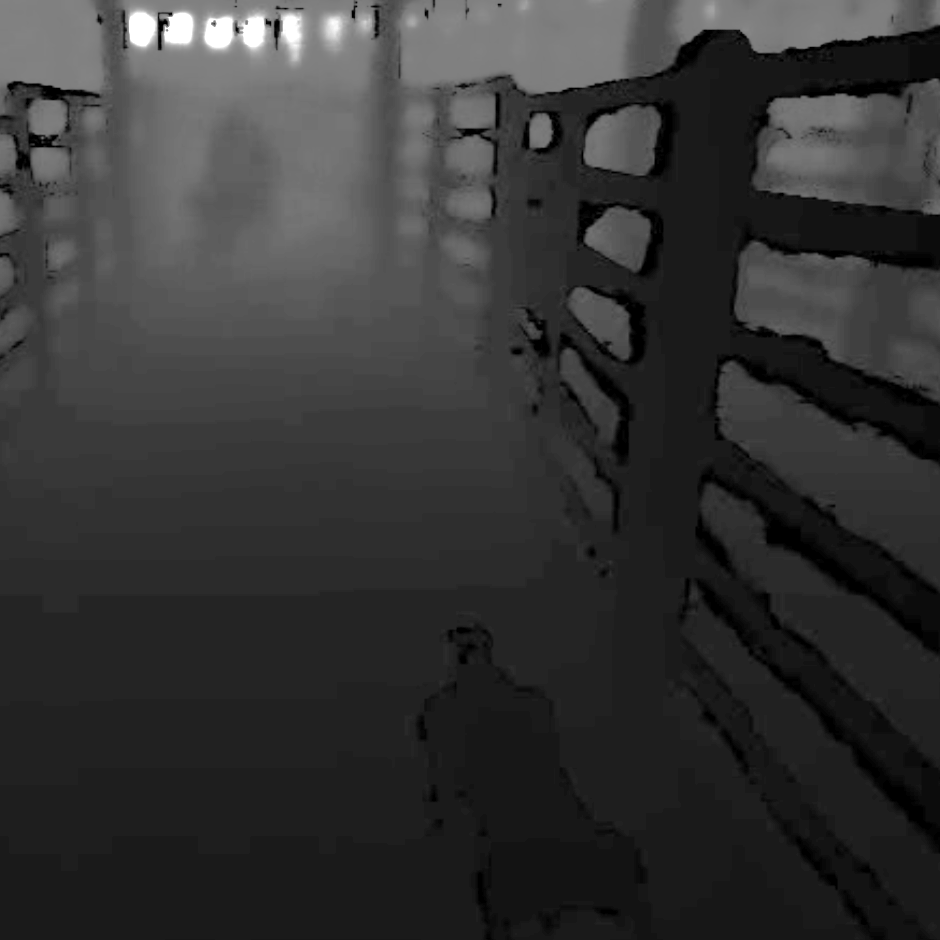}}  &
    {\includegraphics[width=0.2\linewidth]{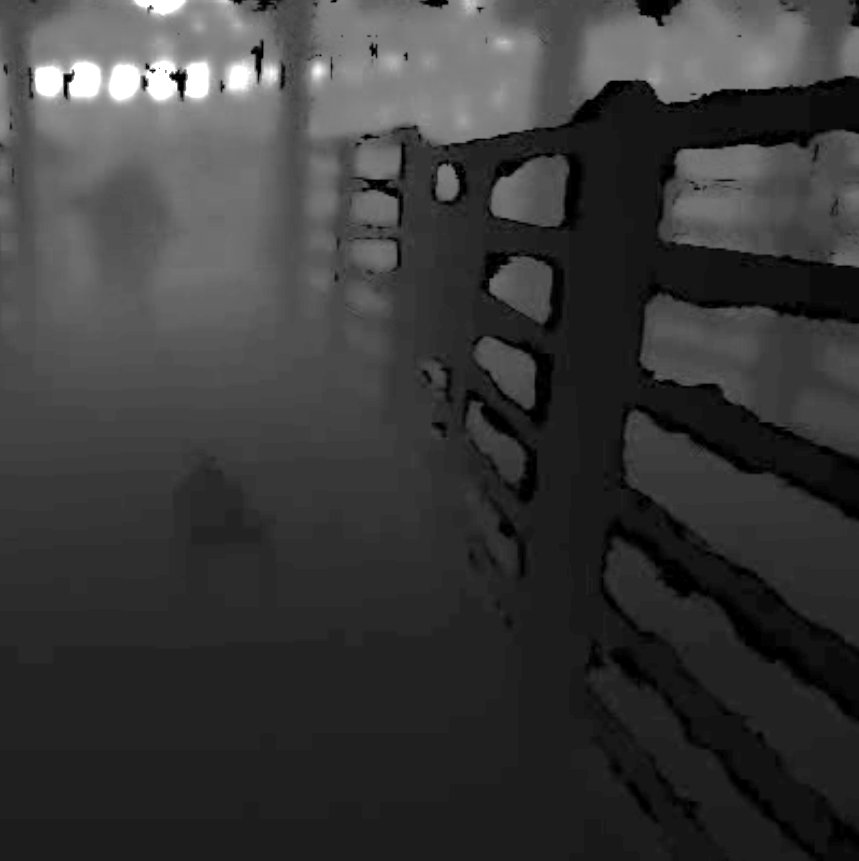}}   &
    {\includegraphics[width=0.2\linewidth]{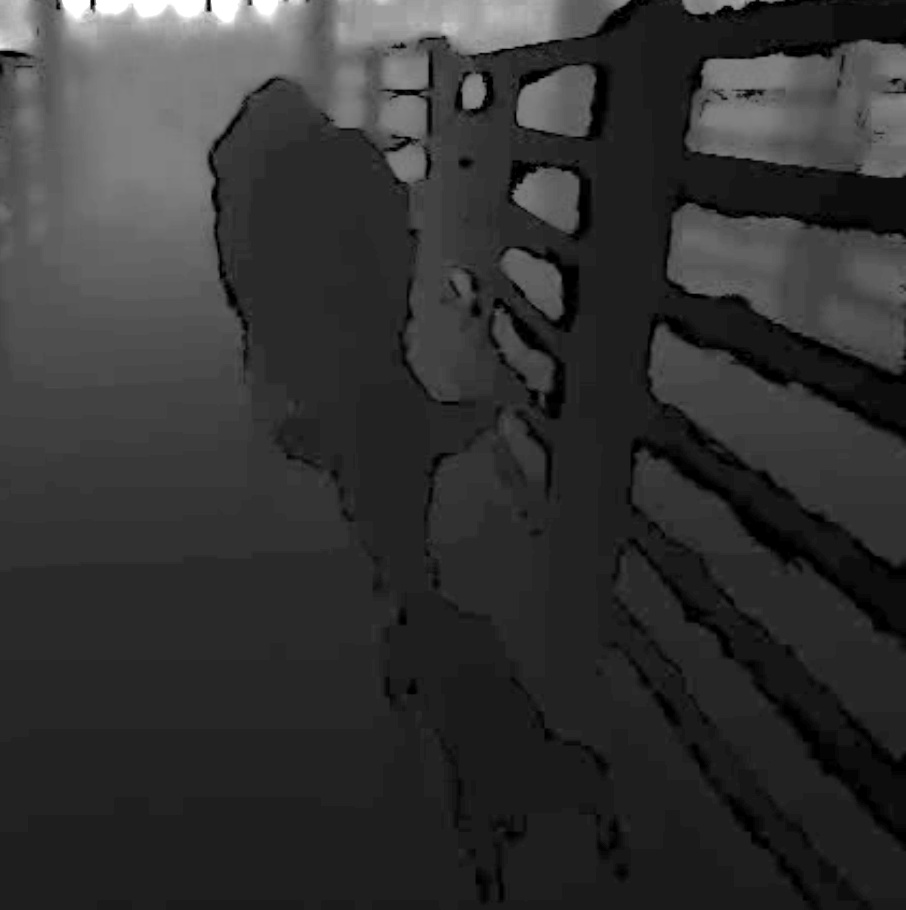}}  &
    {\includegraphics[width=0.2\linewidth]{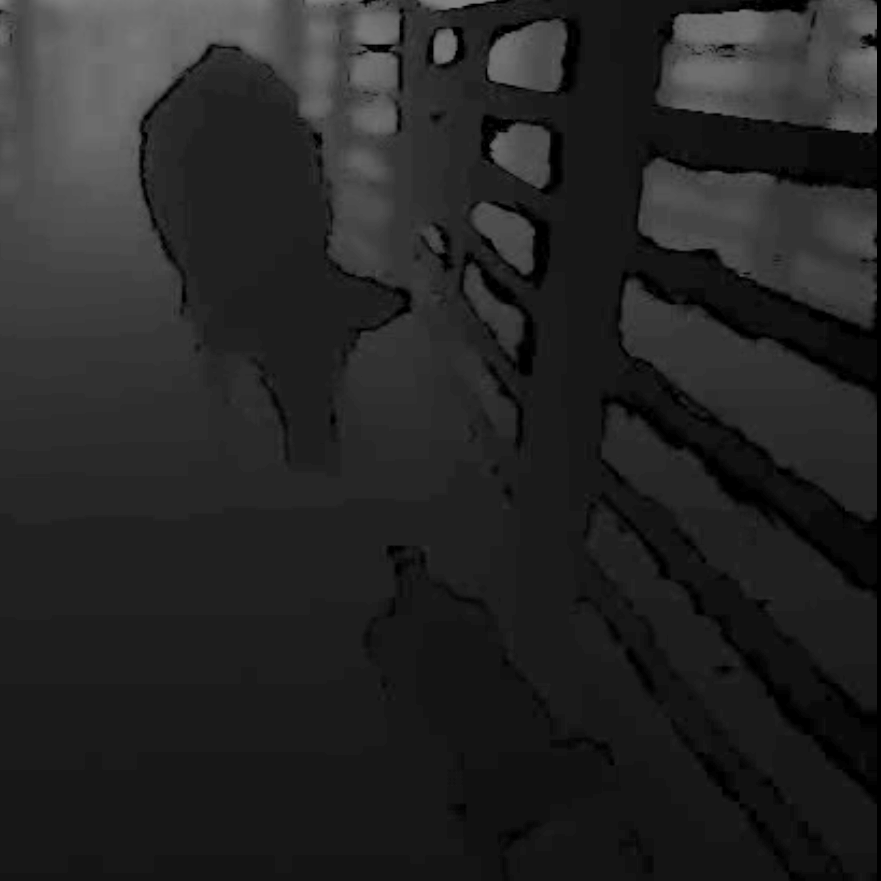}}  \\
    \raisebox{1\height}{\rotatebox[origin=c]{90}{
    \begin{tabular}{c}
    \textit{Unitree GO2}\\
        \textit{1920x1080 RGB} 
        
    \end{tabular}}
} &
    {\includegraphics[width=0.2\linewidth]{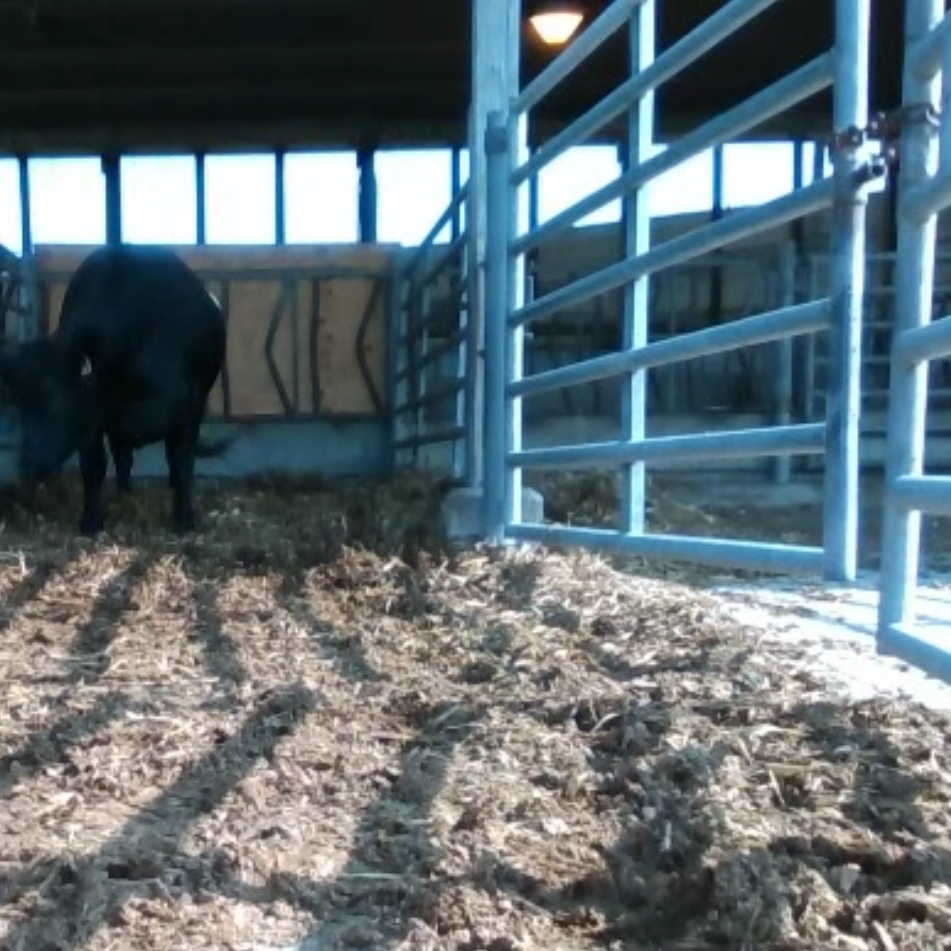}}  &
    {\includegraphics[width=0.2\linewidth]{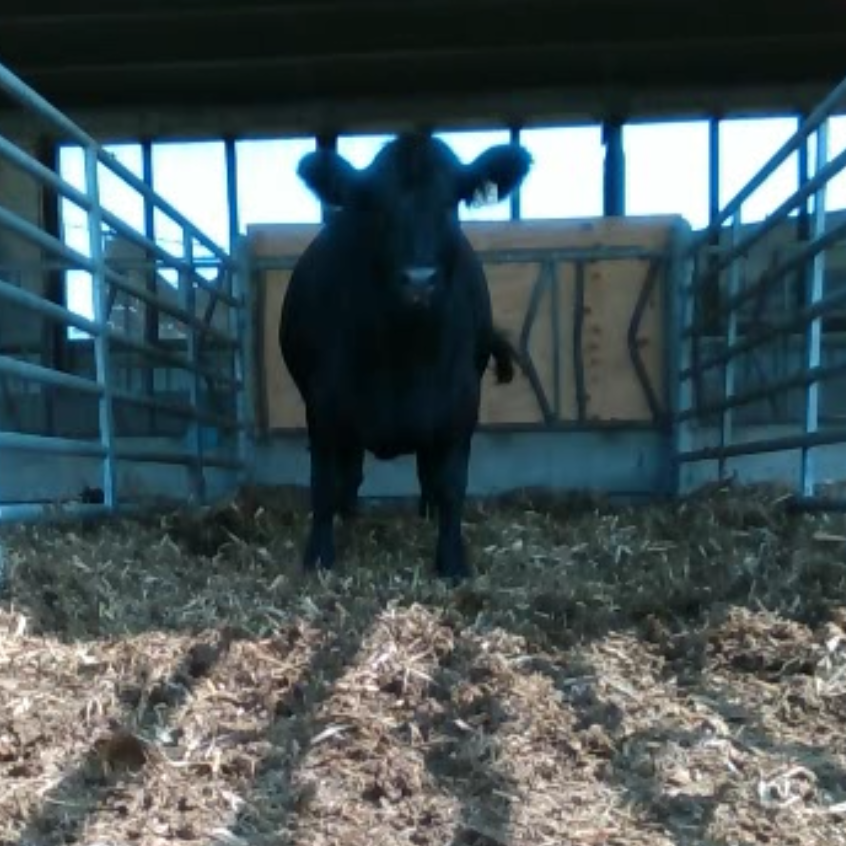}}  &
    {\includegraphics[width=0.2\linewidth]{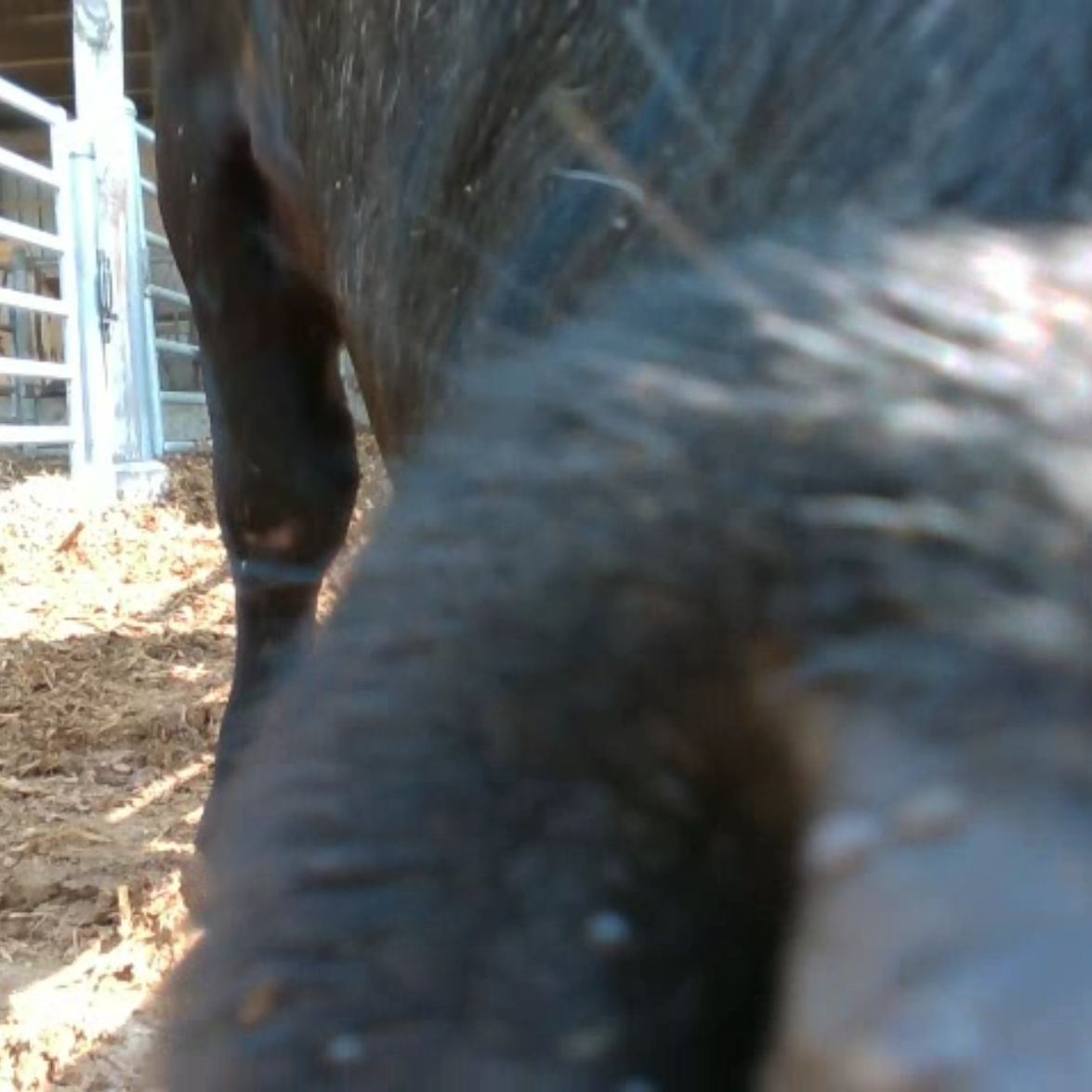}}  &
    {\includegraphics[width=0.2\linewidth]{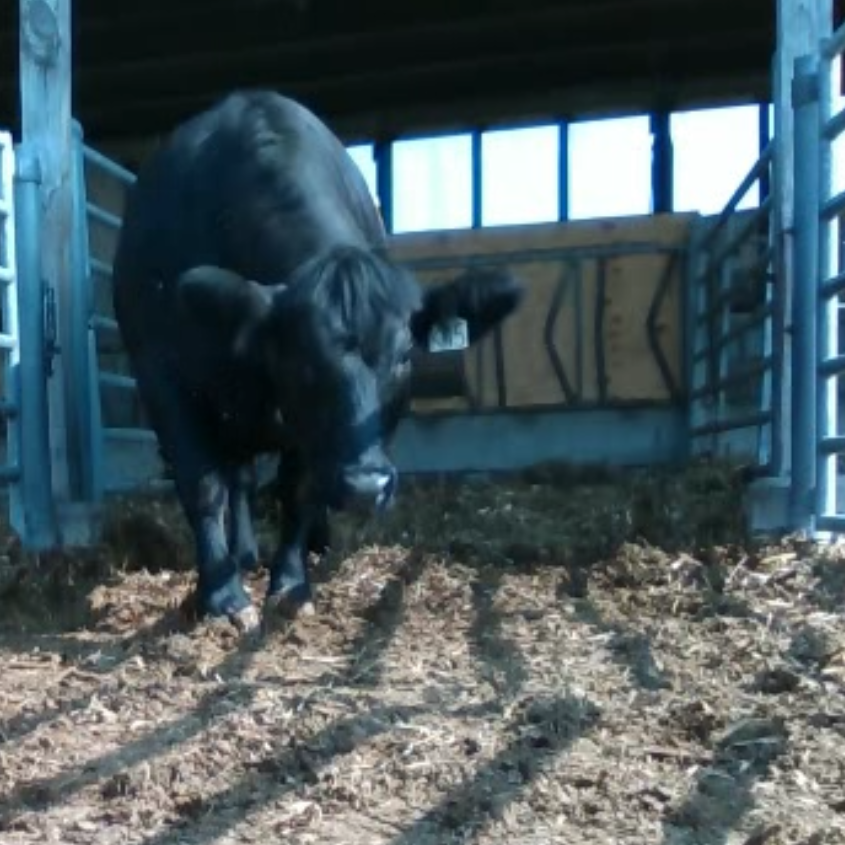}}  \\
\end{tabular}
\caption{Multiple viewpoints captured in the \ari dataset. The rows show camera perspectives (two in the field and one on the robot), while the columns represent task phases: adaptation, approach, negative interaction, and positive interaction.}
\vspace{-0.4 cm}
\label{fig:ari_full_layout}
\end{figure*}

Our experimental protocol comprised four main phases for each interaction session: adaptation, approach, interaction, and retreat (Fig.~\ref{fig:ari_full_layout}). These phases were meticulously designed to systematically evaluate the cow's responses to the robot in a controlled environment.

\textbf{Adaptation Phase:} In this phase, we introduce the robot dog to the cow. This is a crucial stage for familiarizing the animal with the robot, allowing it to become accustomed to the robot's presence before any direct interaction occurs. This phase had two stages: 
\begin{itemize}
    \item \emph{Static Adaptation:} The robot was placed in the cow's line of sight but was kept motionless. This setup enabled the cow to visually perceive the robot without any movements, facilitating initial acclimatization.  
    \item \emph{Dynamic Adaptation:} The robot transitioned from crouching to standing position while remaining in the animal's line of sight. This introduced minimal motion, helping the cow adjust to the robot's ability to move without causing distress. 
\end{itemize}
The phase lasted 30 minutes for each session with a new animal. While initially cautious, no animal showed any signs of aggression during this phase.

\begin{figure*}[!ht]
\vspace{1.4em}
\centering\includegraphics[width=0.98\linewidth]{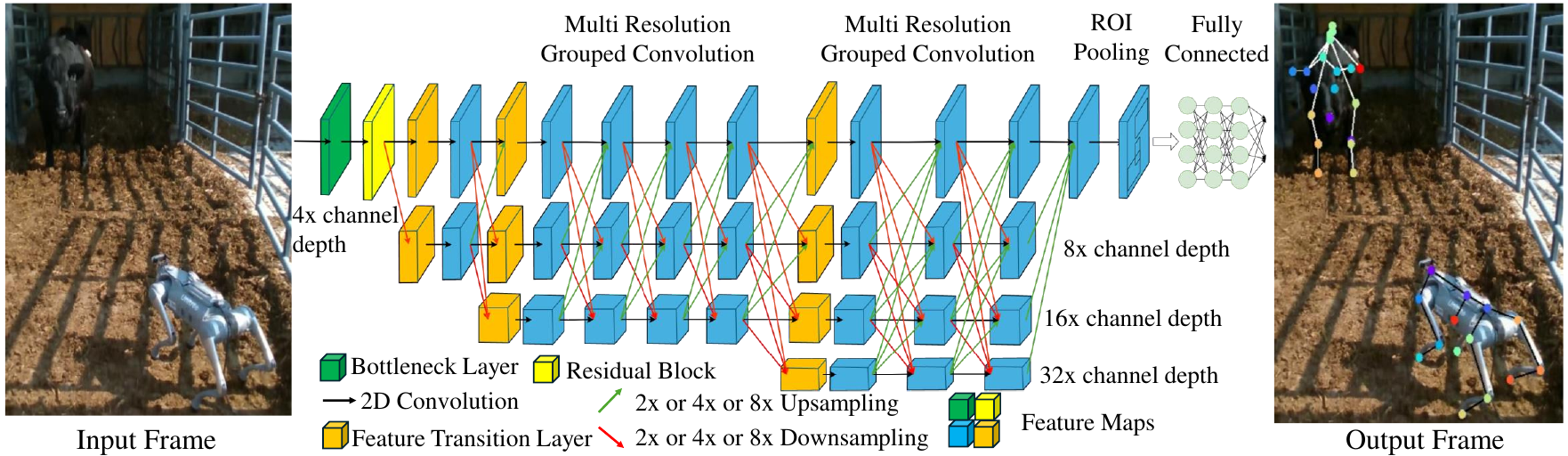}
\caption{The architecture of the body pose estimation model using HRNet.}
\label{fig:hrnet_model}
\vspace{-1.4em}
\end{figure*}

\textbf{Approach Phase:} After adaptation, the robot started a slow approach toward the cow. We conducted the approach at a steady and unhurried pace to avoid startling the animal. The robot paused at predetermined intervals, during which we observed the cow's behavioral cues before deciding to continue. The behavioral cues included:
\begin{itemize}
    \item Inquisitive Orientation: The cow oriented itself toward the robot, indicating curiosity without signs of fear.
    \item Continued Normal Activities: The cow resumed or maintained normal behaviors, such as grazing or resting, suggesting comfort with the robot's presence.
    \item Startle Responses: The cow exhibited signs of startledness, such as sudden movements or a raised head posture.
\end{itemize}
These cues informed our decisions on whether to proceed with the approach, pause for longer periods, or adjust the robot's behavior to minimize stress.

\textbf{Interaction Phase:} Once the robot successfully approached the cow without inducing distress, the interaction phase began. We classified the cow's responses as:
\begin{itemize}
    \item Neutral Interaction: The cow ignored the robot, continuing its previous activities without significant changes in behavior. This response indicated that the cow did not perceive the robot as a threat or an object of interest.
    \item Positive Interaction: The cow displayed interest in the robot through behaviors such as approaching the robot, sniffing, licking, or gently nudging it. These actions suggested a level of comfort, curiosity, and willingness to engage with the robot.
    \item Negative Interaction: The cow exhibited signs of discomfort or distress, including rapid stalling, vocalizations indicative of stress, agitated movements, or defensive postures. Such behaviors signaled that the cow perceived the robot as a threat or an unwelcome presence.
\end{itemize}
We meticulously recorded these interactions, noting the specific behaviors and the context in which they occurred. This detailed annotation allowed us to comprehensively analyze the nuances of ARI.

\textbf{Retreat Phase:} The retreat phase was initiated either after a successful interaction session or immediately upon observing signs of cow distress. During this phase, the robot slowly retreated from the cow while remaining within its line of sight. This gradual retreat prevented startling the cow and allowed us to observe its reactions to the robot’s withdrawal.

\section{Body Pose Estimation} 
Accurate body pose estimation is key to interpreting the emotional and behavioral states of animals, particularly in ARI. %

Leveraging the unique multimodal data provided by the \textit{MBE-ARI} dataset, we introduce a robust neural network-based pose estimation model that tracks 39 keypoints on the cow and 12 on the robot. This extensive keypoint coverage is novel in its ability to capture fine-grained movements, offering far more detailed interaction insights than previously available datasets or models in the field. 

Our approach integrates the High-Resolution Network (HRNet)~\cite{wang2020hrnet} within the Faster R-CNN~\cite{ren2016faster}model to localize keypoints in video frames, as illustrated in Fig.~\ref{fig:hrnet_model}. %
The HRNet backbone processes the regions of interest through parallel multi-resolution streams. This preserves high-resolution spatial details, enabling the Region of Interest (ROI) pooling block to capture fine-grained posture variations that are crucial for interpreting the animal's behavioral intentions.

This system provides precise pose information of the cow’s body throughout the interaction phases, allowing us to correlate specific poses with behavioral outcomes such as fear or curiosity. By refining robot behaviors based on these insights, our framework not only enhances the analysis of interaction dynamics, but also sets a foundation for future advancements in ARI research.

\begin{table}[bh]
    \centering
    \caption{\ari Dataset Summary}
    \label{tab:dataset_overview}
    \begin{tabular}{lcc}
        \toprule
        \textbf{Metric} & \textbf{RGB} & \textbf{Depth} \\
        \midrule
        \textbf{Labeled Images} & 3,000 & 3,000 \\
        \textbf{Total Duration} & 6 hours & 4.8 hours \\
        \textbf{Frame Rate} & 15 fps & 30 fps \\
        \textbf{Number of Cameras} & 5 (4 RealSense, 1 GO2) & 4 RealSense \\
        \textbf{Number of Experiments} & 12 & 12 \\
        \textbf{Data Intervals} & \multicolumn{2}{c}{72 intervals of 5 minutes each} \\
        \textbf{Total Dataset Size} & 1.47\,TB & 1.55\,TB \\
        \bottomrule
    \end{tabular}
\end{table}

\section{Experiments and Results}
In this section, we summarize the dataset. We also show the performance of our novel pose estimation framework on the MBE-ARI dataset, benchmarking it against established models.  

\subsection{\ari Dataset}
The \textit{MBE-ARI} dataset, composed of 3,000 manually labeled video frames, includes physical and activity labels for the robot and the cows. Data were collected over four days, segmented into 5-minute intervals, in 12 different experimental tasks (as summarized in Table~\ref{tab:dataset_overview}). This structure captures a diverse range of behaviors and interaction dynamics, making it an ideal resource for post-processing and training models.

One key strength of this dataset is its ability to capture nuanced animal behaviors that were previously unobserved in similar datasets. For example, our data includes both reserved and curious cow interactions, allowing for richer analysis of how animals respond to robotic presence. Critical interaction behaviors, such as varied locomotion patterns, licking, and smelling, provide valuable insights into animal-robot interaction, which were previously absent in existing datasets. These behaviors offer a detailed understanding of animal curiosity, fear, and adaptation to robots, significantly advancing the field.

Two key findings from the recorded interactions stand out. First, the robot’s slow approach combined with bowing gestures, successfully avoided triggering fear responses in both cows, suggesting a promising adaptation strategy to integrate robots into natural animal habitats. Second, after 30 minutes, the cows resumed normal activities, ignoring the robot, demonstrating that robots can be integrated without causing stress, a breakthrough for deploying robotic systems in environments like precision agriculture. %

\subsection{Pose Estimation Implementation Details}
Our pose estimation framework integrates HRNet with Faster R-CNN and was implemented using the DeepLabCut markerless pose estimation tool~\cite{Mathis2018}. We trained the model on the multimodal video streams of the MBE-ARI dataset, augmenting the data with transformations like rotation, shifts, and brightness changes to improve robustness across real-world conditions. This diversity of augmentation, achieved using the Albumentations library~\cite{buslaev2020albumentations}, ensured the model could generalize across environments and animal behaviors.

The HRNet backbone was pre-trained on the Super Animal Quadruped datasets~\cite{super}, which integrates multiple animal pose datasets including AP-10, AnimalPose, iRodent, Horse-10, and StanfordDogs. This diverse pre-training facilitates robust feature extraction across various animal species and poses. The model was trained using PyTorch, with periodic updates and checkpoints to monitor progress and ensure model stability. Specifically, training metrics were logged every 1,000 iterations, and model checkpoints were saved every 50 epochs, with training run for a maximum of 200 epochs. This approach allowed for careful tracking of the model’s learning process and prevented overfitting. After training, the model was tested across the entire \ari dataset for keypoint detection, with manual refinement used to correct for outliers in the predictions, ensuring high precision.

\begin{figure}[!ht]
\centering\includegraphics[width=1.0\linewidth]{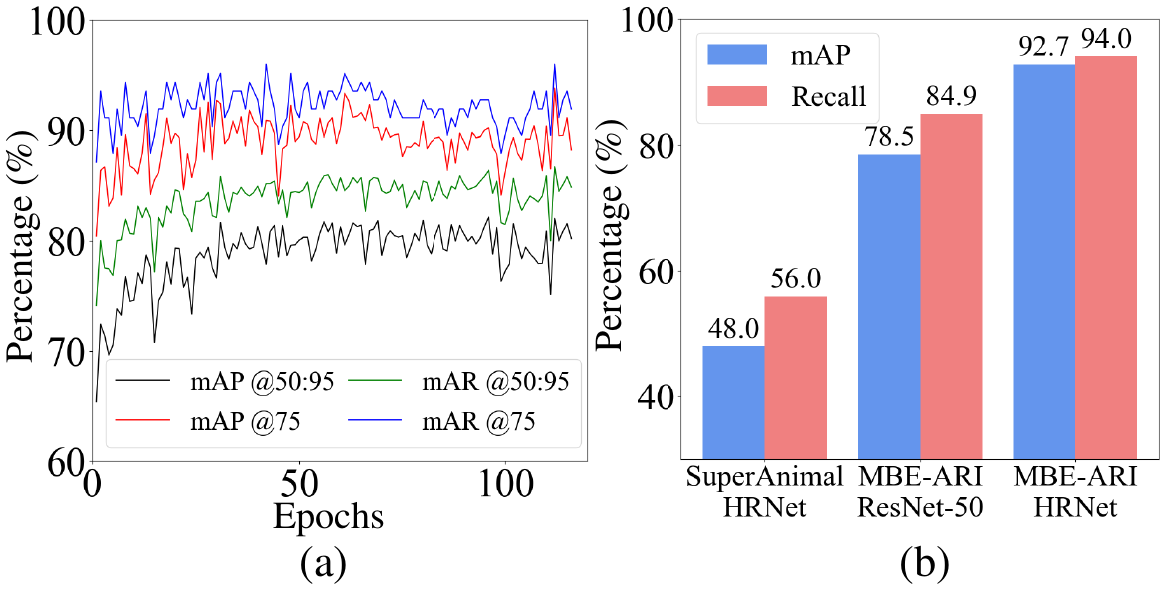}
\caption{(a) Training performance of our pose estimation model with the HRNet architecture, showing mAP and mAR at IoU thresholds of 50:95 and 75. (b) Comparison of mAP and mAR performance between our framework, the Super Animal HRNet, and our framework with a ResNet-50 backbone.}
\label{fig:results}
\vspace{-0.3cm}
\end{figure}

\subsection{Evaluation Metrics}
We evaluated the performance of our framework on the \ari dataset using standard keypoint detection metrics, specifically, mean average precision (mAP) and mean average recall (mAR) at various intersections over union (IoU) thresholds. The reported metrics include mAP@75 (mean average precision at an IoU threshold of 75\%) and mAP@50:95 (mean average precision averaged over IoU thresholds from 50\% to 95\%). Similarly, we measured mAR@75 and mAR@50:95 to assess the model's ability to correctly identify keypoints under different levels of localization strictness. These metrics provide a comprehensive evaluation of the model's precision and recall, reflecting model’s ability to accurately localize keypoints across different scenarios.

\subsection{Body Pose Estimation Results}

The evaluation results for different mAP and mAR threholds are shown in Fig.~\ref{fig:results}(a).
During model training, we observed that at strict IoU thresholds -- specifically 75\% and the range from 50\% to 95\% --the mAP stabilized around 93\%, indicating that the model performs well even in challenging cases, although with some variance. The mAR followed a similar trend, with mAR@75 and mAR@50:95 stabilizing around 94\%. These results highlight the model's ability to balance precision and recall, achieving strong performance across varying levels of detection difficulty.

In our evaluations, for the pose estimation task, our proposed framework achieved an mAP of 92.7\% on the \ari dataset, demonstrating superior performance in accurately localizing keypoints specific to animal poses during interactions with the robot. For comparison, an HRNet model pre-trained on the Super Animal Quadruped dataset achieved only 48\% mAP on the same task, and a simpler ResNet-50 model attained 78.5\% mAP, as shown in Fig.~\ref{fig:results}(b).

Moreover, in prior work related to animal pose estimation~\cite{super}, the same model architecture with an HRNet backbone achieved an mAP of 70.54\% on the AP10 dataset and 58.85\% on the iRodent dataset. The significant improvement in our results underscores the effectiveness of the \ari dataset in capturing the nuanced poses of cows, which are critical to understanding their behavioral responses.

\section{Conclusion and Discussion}
In this work, we have made significant strides in advancing the field of ARI through the introduction of the \textit{MBE-ARI} dataset and a pose estimation framework. This multimodal dataset captures previously unobserved, nuanced interactions between a legged robot and sentient mammals (cows), providing synchronized, labeled data that enables in-depth analysis of behavioral cues such as posture, locomotion, and eye movements. By offering a rich set of interaction dynamics, our dataset addresses a critical gap in the study of ARI, where complex animal behaviors had not been adequately captured in previous datasets.

Our experimental findings reveal two critical insights. First, the use of slow approaches combined with bowing gestures by the robot successfully prevented fear responses in the cows, suggesting a promising strategy for integrating robots into natural environments without inducing stress. Second, the cows' gradual acclimatization to the robot--where they resumed normal activities after 30 minutes--demonstrates that robots can seamlessly integrate into animal habitats, a breakthrough with implications for autonomous systems in natural environments.

Our pose estimation system, which achieved an impressive mAP of 92.7\%, significantly outperforms existing models in tracking subtle animal behaviors. The system’s ability to precisely localize 39 keypoints across varied interactions sets a new benchmark for ARI research and enables the development of adaptive robotic systems capable of real-time responses to animal cues.

The \textit{MBE-ARI} dataset and our pose estimation framework lay the groundwork for future advancements in robotic perception, animal welfare, and environmental conservation, opening the door to more intelligent and adaptive robotic systems that can safely operate alongside sentient animals in real-world settings. In the future, we will leverage this dataset to develop dynamic path planning and autonomous control algorithms that can enable robots to navigate and interact seamlessly in complex natural environments. 

\balance{
\bibliographystyle{IEEEtran}
\bibliography{my_bib}}

\begin{thebibliography}{10}
\providecommand{\url}[1]{#1}
\csname url@samestyle\endcsname
\providecommand{\newblock}{\relax}
\providecommand{\bibinfo}[2]{#2}
\providecommand{\BIBentrySTDinterwordspacing}{\spaceskip=0pt\relax}
\providecommand{\BIBentryALTinterwordstretchfactor}{4}
\providecommand{\BIBentryALTinterwordspacing}{\spaceskip=\fontdimen2\font plus
\BIBentryALTinterwordstretchfactor\fontdimen3\font minus \fontdimen4\font\relax}
\providecommand{\BIBforeignlanguage}[2]{{%
\expandafter\ifx\csname l@#1\endcsname\relax
\typeout{** WARNING: IEEEtran.bst: No hyphenation pattern has been}%
\typeout{** loaded for the language `#1'. Using the pattern for}%
\typeout{** the default language instead.}%
\else
\language=\csname l@#1\endcsname
\fi
#2}}
\providecommand{\BIBdecl}{\relax}
\BIBdecl

\bibitem{ruotolo2021grasping}
W.~Ruotolo, D.~Brouwer, and M.~R. Cutkosky, ``From grasping to manipulation with gecko-inspired adhesives on a multifinger gripper,'' \emph{Science Robotics}, vol.~6, no.~61, p. eabi9773, 2021.

\bibitem{legged_loco}
D.~W. Haldane, K.~C. Peterson, F.~L. Garcia~Bermudez, and R.~S. Fearing, ``Animal-inspired design and aerodynamic stabilization of a hexapedal millirobot,'' in \emph{2013 IEEE International Conference on Robotics and Automation}, 2013, pp. 3279--3286.

\bibitem{baisch2014high}
A.~T. Baisch, O.~Ozcan, B.~Goldberg, D.~Ithier, and R.~J. Wood, ``High speed locomotion for a quadrupedal microrobot,'' \emph{The International Journal of Robotics Research}, vol.~33, no.~8, pp. 1063--1082, 2014.

\bibitem{agile_robo_fish}
S.~Zhang, Y.~Qian, P.~Liao, F.~Qin, and J.~Yang, ``Design and control of an agile robotic fish with integrative biomimetic mechanisms,'' \emph{IEEE/ASME Transactions on Mechatronics}, vol.~21, no.~4, pp. 1846--1857, 2016.

\bibitem{robotic_hand}
Z.~Xu and E.~Todorov, ``Design of a highly biomimetic anthropomorphic robotic hand towards artificial limb regeneration,'' in \emph{2016 IEEE International Conference on Robotics and Automation (ICRA)}, 2016, pp. 3485--3492.

\bibitem{zebra_fish}
F.~Bonnet and F.~Mondada, ``Shoaling with fish: using miniature robotic agents to close the interaction loop with groups of zebrafish danio rerio,'' \emph{Springer Tracts in Advanced Robotics}, 2019.

\bibitem{zebra_fish_12}
F.~Bonnet, P.~Rétornaz, J.~Halloy, A.~Gribovskiy, and F.~Mondada, ``Development of a mobile robot to study the collective behavior of zebrafish,'' \emph{2012 4th IEEE RAS EMBS International Conference on Biomedical Robotics and Biomechatronics (BioRob)}, 2012.

\bibitem{rat_behave_telensky}
P.~Telenský, J.~Svoboda, K.~Blahna, J.~Bures, S.~Kubík, and A.~Stuchlı́k, ``Functional inactivation of the rat hippocampus disrupts avoidance of a moving object,'' \emph{Proceedings of the National Academy of Sciences}, vol. 108, pp. 5414--5418, 2011.

\bibitem{rat_ahuja}
N.~Ahuja, V.~Lobellová, A.~Stuchlik, and E.~Kelemen, ``Navigation in a space with moving objects: rats can avoid specific locations defined with respect to a moving robot,'' \emph{Frontiers in Behavioral Neuroscience}, vol.~14, 2020.

\bibitem{speech_hri}
M.~Podpora, A.~Gardecki, R.~Beniak, B.~Klin, J.~Vicario, and A.~Kawala-Sterniuk, ``Human interaction smart subsystem—extending speech-based human-robot interaction systems with an implementation of external smart sensors,'' \emph{Sensors}, vol.~20, p. 2376, 2020.

\bibitem{language_social_robotics}
M.~Foster, R.~Alami, O.~Gestranius, O.~Lemon, M.~Niemelä, J.~Odobez, and A.~Pandey, ``The mummer project: engaging human-robot interaction in real-world public spaces,'' in \emph{Social Robotics: 8th International Conference}, 2016, pp. 753--763.

\bibitem{lambert2019positive}
H.~Lambert and G.~Carder, ``Positive and negative emotions in dairy cows: Can ear postures be used as a measure?'' \emph{Behavioural processes}, vol. 158, pp. 172--180, 2019.

\bibitem{kaur2021future}
U.~Kaur, R.~M. Voyles, and S.~Donkin, ``Future of animal welfare-technological innovations for individualized animal care,'' \emph{Improving animal welfare}, vol. 570, 2021.

\bibitem{kaur2023invited}
U.~Kaur, V.~M. Malacco, H.~Bai, T.~P. Price, A.~Datta, L.~Xin, S.~Sen, R.~A. Nawrocki, G.~Chiu, S.~Sundaram \emph{et~al.}, ``Invited review: integration of technologies and systems for precision animal agriculture—a case study on precision dairy farming,'' \emph{Journal of Animal Science}, vol. 101, p. skad206, 2023.

\bibitem{robot_gaze}
M.~Zheng, A.~Moon, E.~Croft, and M.~Meng, ``Impacts of robot head gaze on robot-to-human handovers,'' \emph{International Journal of Social Robotics}, vol.~7, pp. 783--798, 2015.

\bibitem{olaronke_robot_cues}
I.~Olaronke, O.~Oluwaseun, and R.~Ikono, ``State of the art: a study of human-robot interaction in healthcare,'' \emph{International Journal of Information Engineering and Electronic Business}, vol.~9, pp. 43--55, 2017.

\bibitem{mathis2018deeplabcut}
A.~Mathis, P.~Mamidanna, K.~M. Cury, T.~Abe, V.~N. Murthy, M.~W. Mathis, and M.~Bethge, ``Deeplabcut: markerless pose estimation of user-defined body parts with deep learning,'' \emph{Nature neuroscience}, vol.~21, no.~9, pp. 1281--1289, 2018.

\bibitem{mathis2020deep}
M.~W. Mathis and A.~Mathis, ``Deep learning tools for the measurement of animal behavior in neuroscience,'' \emph{Current opinion in neurobiology}, vol.~60, pp. 1--11, 2020.

\bibitem{lauer2022multi}
J.~Lauer, M.~Zhou, S.~Ye, W.~Menegas, S.~Schneider, T.~Nath, M.~M. Rahman, V.~Di~Santo, D.~Soberanes, G.~Feng \emph{et~al.}, ``Multi-animal pose estimation, identification and tracking with deeplabcut,'' \emph{Nature Methods}, vol.~19, no.~4, pp. 496--504, 2022.

\bibitem{wang2020hrnet}
J.~Wang, K.~Sun, T.~Cheng, B.~Jiang, C.~Deng, Y.~Zhao, D.~Liu, Y.~Mu, M.~Tan, X.~Wang \emph{et~al.}, ``Deep high-resolution representation learning for visual recognition,'' \emph{IEEE transactions on pattern analysis and machine intelligence}, vol.~43, no.~10, pp. 3349--3364, 2020.

\bibitem{ren2016faster}
S.~Ren, K.~He, R.~Girshick, and J.~Sun, ``Faster r-cnn: Towards real-time object detection with region proposal networks,'' \emph{IEEE transactions on pattern analysis and machine intelligence}, vol.~39, no.~6, pp. 1137--1149, 2016.

\bibitem{Mathis2018}
A.~Mathis, P.~Mamidanna, K.~M. Cury, T.~Abe, V.~N. Murthy, M.~W. Mathis, and M.~Bethge, ``Deeplabcut: markerless pose estimation of user-defined body parts with deep learning,'' \emph{Nature Neuroscience}, vol.~21, pp. 1281--1289, 9 2018.

\bibitem{buslaev2020albumentations}
A.~Buslaev, V.~I. Iglovikov, E.~Khvedchenya, A.~Parinov, M.~Druzhinin, and A.~A. Kalinin, ``Albumentations: fast and flexible image augmentations,'' \emph{Information}, vol.~11, no.~2, p. 125, 2020.

\bibitem{super}
S.~Ye, A.~Filippova, J.~Lauer, S.~Schneider, M.~Vidal, T.~Qiu, A.~Mathis, and M.~W. Mathis, ``Superanimal pretrained pose estimation models for behavioral analysis,'' \emph{Nature Communications}, 2024.

\end{thebibliography}
\end{document}